\documentclass[10pt,twocolumn,letterpaper]{article}

\usepackage{cvpr}              
\definecolor{cvprblue}{rgb}{0.21,0.49,0.74}
\usepackage[pagebackref,breaklinks,colorlinks,allcolors=cvprblue]{hyperref}

\usepackage[safe]{tipa}
\usepackage{tabularx, makecell, booktabs}
\usepackage{multirow}
\newcommand\nonumberfootnote[1]{%
  \begingroup%
  \let\thefootnote\relax%
  \footnotetext{#1}%
  \addtocounter{footnote}{-1}%
  \endgroup%
}

\def\paperID{4594} 
\def\confName{CVPR}
\def\confYear{2026}

\title{TAVID: Text-Driven Audio-Visual Interactive Dialogue Generation}

\author{Ji-Hoon Kim\(^{1, 2*}\) \quad Junseok Ahn\(^{1*}\) \quad Doyeop Kwak\(^1\) \quad Joon Son Chung\(^1\) \quad Shinji Watanabe\(^2\) \\ 
$^1$Korea Advanced Institute of Science and Technology \quad $^2$Carnegie Mellon University\\
{\tt\small \{jh.kim, junseok.ahn\}@kaist.ac.kr}
}

\begin{document}
\twocolumn[{
\maketitle
\vspace{-7mm}
\begin{center}
    \captionsetup{type=figure}
    \includegraphics[width=0.94\textwidth]{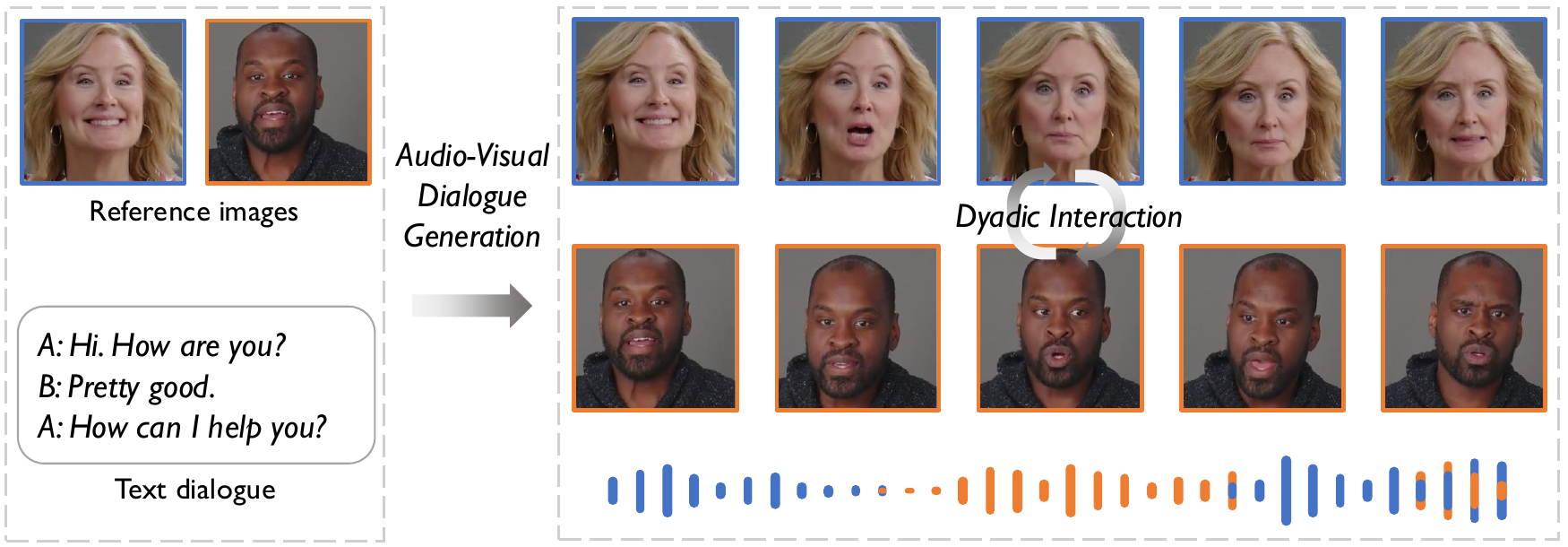}
    \captionof{figure}{Overview of TAVID framework. 
    Given a text dialogue and reference images, TAVID simultaneously produces interactive videos and conversational speech with natural turn-taking, accurate synchronization and expressive facial dynamics.}
    \label{fig:teaser}
\end{center}
}]
\nonumberfootnote{\hspace{-3mm}$^{*}$Equal contribution.}
\nonumberfootnote{\hspace{-3mm}Project Page: \url{https://mm.kaist.ac.kr/projects/TAVID}}
\begin{abstract}
The objective of this paper is to jointly synthesize interactive videos and conversational speech from text and reference images. 
With the ultimate goal of building human-like conversational systems, recent studies have explored talking or listening head generation as well as conversational speech generation.
However, these works are typically studied in isolation, overlooking the multimodal nature of human conversation, which involves tightly coupled audio-visual interactions. 
In this paper, we introduce TAVID, a unified framework that generates both interactive faces and conversational speech in a synchronized manner. 
TAVID integrates face and speech generation pipelines through two cross-modal mappers (i.e., a motion mapper and a speaker mapper), which enable bidirectional exchange of complementary information between the audio and visual modalities.
We evaluate our system across four dimensions: talking face realism, listening head responsiveness, dyadic interaction fluency, and speech quality. 
Extensive experiments demonstrate the effectiveness of our approach across all these aspects.
\end{abstract}    
\section{Introduction}
\label{sec:intro}
Have you ever imagined having a natural conversation with an AI? 
Indeed, there have been numerous efforts to build systems capable of fluent communication, reflecting growing demand in areas such as AI tutoring, virtual companionship, and social robotics.
However, such systems have predominantly been limited to a single modality, such as text~\cite{OpenAI2023GPT4TR,grattafiori2024llama,yang2025qwen3} or speech~\cite{nguyen2023generative:dGSLM,defossez2024moshi,veluri2024beyond}.
In contrast, human communication is inherently multimodal, combining linguistic content with vocal and visual cues that enrich nuance, emotion, and intent~\cite{mehrabian1971silent}.
Therefore, to create truly immersive and realistic interactions between human and AI, it is crucial to integrate information across multiple modalities, rather than relying on text or speech alone.

With the ultimate goal of building human-like conversational agents, prior work has largely been fragmented into independent lines of research, including talking head generation and listening head generation.
Talking head generation focuses on synthesizing a speaker's lip and head motions driven by an audio~\cite{prajwal2020lip,jang2023s,xu2024vasa} or a text signal~\cite{choi2024text, jang2024faces,diao2025ft2tf}. 
In parallel, listening head generation aims to produce a listener's facial feedback in response to the speaker's acoustic and visual behaviors~\cite{zhou2022responsive:VICO,ng2022learning,song2023emotional,liu2024customlistener}. 
Although these two tasks have succeeded in animating natural faces, they focus solely on one-sided communication, overlooking the dyadic nature of human conversation.

To model dyadic communication, recent studies have explored interactive head generation. 
Early works~\cite{tran2024dim,wang2024disentangling,zhou2025interactive} rely on manually defined role switchers to alternate between speaking and listening states, which often lead to unnatural transitions. 
To address this issue, INFP~\cite{zhu2024infp} proposes an interactive motion guider that automatically determines the state using dyadic motion representations driven by dual-track audio.
Recently, ARIG~\cite{guo2025arig} further improves interaction realism and generation quality by incorporating long-range contextual cues from both audio and visual modalities.
Despite these advances towards conversational agents, existing methods rely on pre-recorded audio to produce facial videos, making them incapable of creating a new speech content.
Although a common workaround is to construct cascaded systems integrating text-to-speech (TTS) networks, this approach inevitably suffers from error accumulation and additional speaker modeling such as acoustic prompting~\cite{jang2024faces,wang2025omnitalker}.

In this paper, we propose TAVID, a unified framework for \textbf{T}ext-driven \textbf{A}udio-\textbf{V}isual \textbf{I}nteractive \textbf{D}ialogue generation. 
As illustrated in \cref{fig:teaser}, TAVID jointly generates conversational speech and interactive videos from a text dialogue and reference images, enabling flexible content creation and automatic speaker modeling. 
To this end, TAVID integrates video and speech generation pipelines with two cross-modal mappers--the \textit{Motion Mapper} and the \textit{Speaker Mapper}--which capture mutually complementary information across the two streams. 
The \textit{Motion Mapper} converts text dialogues into dyadic motion features that dynamically alternate between speaking and listening states. 
To ensure accurate and interactive motion, we analyze several architectural schemes and adopt a joint self-attention mechanism that facilitates bi-directional information exchange between speakers. 
Meanwhile, the \textit{Speaker Mapper} ensures that the synthesized voice aligns with the visual persona by producing speaker features consistent with the visual identity. 

Our main contributions are summarized as follows:
\begin{itemize}
    \item We propose TAVID, a unified framework for text-driven interactive dialogue generation.
    To our knowledge, this is the first attempt to jointly generate both interactive video and conversational speech from text inputs, enabling flexible control over spoken content while ensuring visually consistent speaker characteristics.
    \item We design two novel cross-modal mappers, the Motion Mapper and Speaker Mapper, that predict complementary features across audio and visual streams, facilitating a synergistic audio-visual interaction.
    \item Our method generates not only realistic interactive videos but also high-quality conversational speech, demonstrating its effectiveness across multiple aspects, including talking face realism, listener responsiveness, dyadic interaction fluency, and speech naturalness.
\end{itemize}

\section{Related Works}

\begin{figure*}[t]
    \centering
    \includegraphics[width=\textwidth]{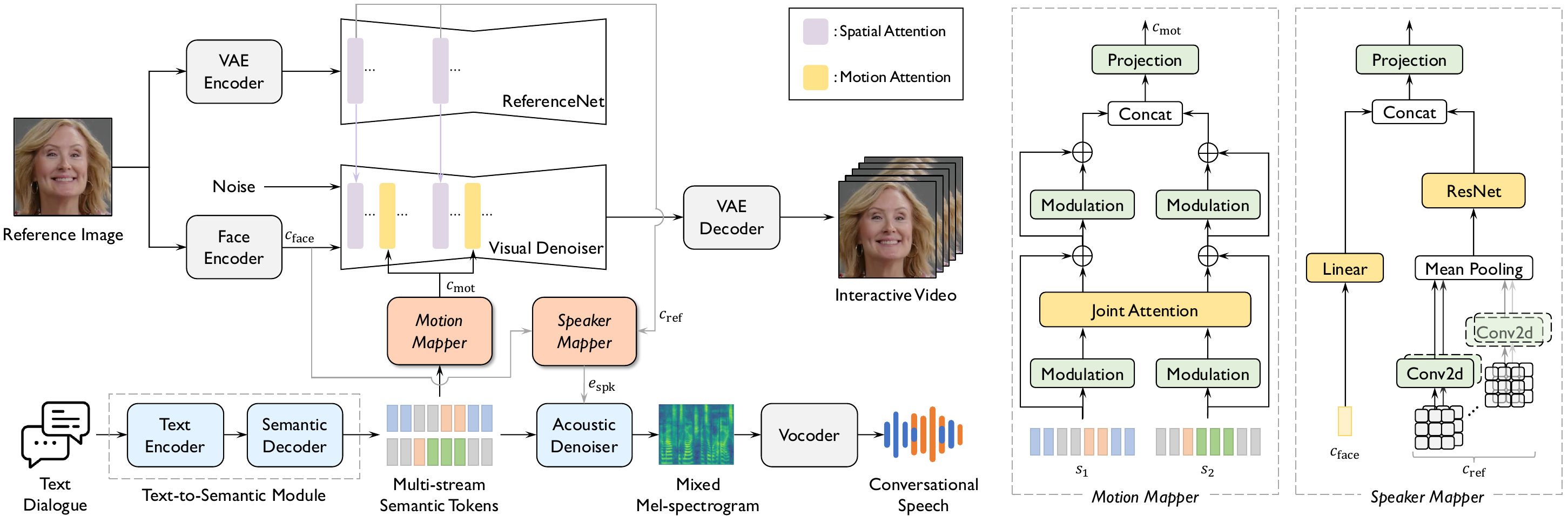}
    \caption{
    The overall architecture of TAVID.
    Given a text dialogue and a reference image, our approach generates both interactive video and conversational speech, guided by two cross-modal mappers.
    The \textit{Motion Mapper} predicts interactive motions from multi-stream semantic tokens, while the \textit{Speaker Mapper} models vocal characteristics from the reference image.
    The detailed architectures of the cross-modal mappers are shown on the right side.
    }
    \vspace{-1mm}
    \label{fig:model}
\end{figure*}

\subsection{Single Role Head Generation}
Single role head generation refers to either talking head or listening head generation.
Talking head generation, which aims to produce speakers' facial video with accurate lip and head movements, can be broadly divided into two categories.
One line of work utilizes audio signals to animate a reference image, typically based on generative adversarial networks~\cite{prajwal2020lip,park2022synctalkface,zhang2023sadtalker} or diffusion networks~\cite{cui2024hallo2,chen2025echomimic,cui2024hallo3}. 
These approaches assume that an input speech signal is already available, either as a real recording of a human voice or generated by a pre-trained text-to-speech (TTS) networks.
The other line utilizes a text instead of audio ~\cite{jang2024faces,diao2025ft2tf,chatziagapi2025av,wang2025omnitalker}.
These works jointly synthesize speech and talking head videos from text inputs, enabling flexible control over linguistic content while avoiding the latency and error accumulation associated with TTS-cascaded systems.

In parallel, listening head generation aims to produce facial videos of a listener exhibiting natural responsive behaviors, such as attentive gaze and subtle head movements.
As one of the earliest efforts in this direction, RLHG~\cite{zhou2022responsive:VICO} introduces the ViCo dataset, establishing a standard benchmark for responsive listener generation.
Following to this, this field has evolved to model the non-deterministic nature of listener feedback by generating diverse~\cite{ng2022learning}, emotionally controllable~\cite{song2023emotional}, and text-prompted responses~\cite{liu2024customlistener}.
Although these single role methods have demonstrated impressive performance, the explicit separation between speaker and listener roles constrains the synthesis of fluid and natural dyadic interactions.
In contrast, the proposed approach jointly models both roles, enabling coherent and contextually adaptive interactions prompted by text inputs.

\subsection{Interactive Head Generation}
To capture the dyadic nature of human conversations, there has been a growing line of research on interactive head generation.
This task aims to produce facial videos that seamlessly switch between speaking and listening roles within multi-turn conversations.
Early works~\cite{tran2024dim,wang2024disentangling,zhou2025interactive} model the speaking and listening roles in separate branches and manually assign role states to their respective branches.
However, such manual role assignment often leads to unnatural transitions between different conversational states.
Furthermore, this paradigm fails to comprehensively capture all possible states in dyadic conversations, such as when both participants speak simultaneously or remain silent.
Recently, INFP~\cite{zhu2024infp} introduces a key breakthrough for this task.
By introducing an interactive motion guider, they generate mixed speaking-listening motions that dynamically alternate between speaking and listening states without requiring explicit role assignment.
Following this, ARIG~\cite{guo2025arig} further enhances interaction realism and visual quality by incorporating long-range contextual information from audio–visual modalities.
While these efforts have achieved promising results in modeling conversational dynamics, they still rely on pre-recorded speech, inheriting the fundamental limitations of audio-driven talking head generation.
Different from prior works, we propose a text-driven interactive head generation system, which not only enables flexible speech content editing but also automatically aligns voice identity with visual identity.

\subsection{Conversational Speech Generation}
Conversational speech generation aims to produce coherent multi-turn speech while capturing key characteristics of human conversation, such as natural turn-taking and back-channeling. 
As a pioneering work, dGSLM~\cite{nguyen2023generative:dGSLM} models the turn-taking dynamics of conversational speech through a dual-tower Transformer, serving as the foundation for subsequent works such as CHATS~\cite{mitsui2023towards:CHATS} and SLIDE~\cite{lu2025slide}.
Meanwhile, several attempts have been made to generate conversational speech from text dialogues, most of which are based on autoregressive architectures~\cite{ju2025mooncast, deepmind-audio-generation, peng2025vibevoice}. 
These models predict either semantic tokens~\cite{zhang2024covomix} or acoustic codec tokens~\cite{nari-labs-dia,darefsky2024parakeet} from text input.
More recently, non-autoregressive architectures have also been explored to address the inherent limitations of autoregressive approaches~\cite{zhang2025covomix2, zhu2025zipvoice}.
Although these models have succeeded in synthesizing naturalistic conversational speech, they fundamentally lack the visual dimension required for holistic human communication.
Our approach, however, generates interactive videos alongside naturalistic speech, embodying the multimodal nature of human communication.

\begin{figure}[!t]
    \centering
    \subfloat[][Concatenation]{
        \includegraphics[height=25.5mm]{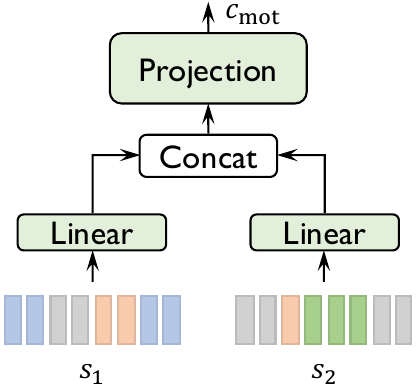}
        \label{fig:motionmapper:concat}
    }
    \subfloat[][Dual Attention]{
        \includegraphics[height=25.5mm]{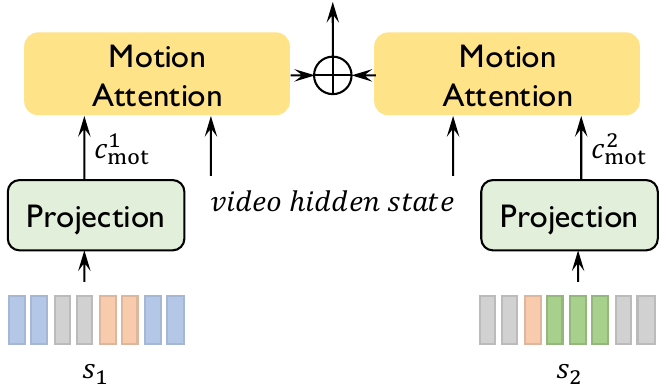}
        \label{fig:motionmapper:dual}
    }
    \caption{
    Analysis of integration strategies for the \textit{Motion Mapper}.
    $c_\text{mot}$ denotes the interactive motion features extracted from multi-stream semantic tokens, whereas $c_\text{mot}^{1}$ and $c_\text{mot}^{2}$ correspond to motion features specific to each individual stream.
    }
    \label{fig:motionmapper}
\end{figure}
\section{TAVID}
\subsection{Network Architecture}

As depicted in \cref{fig:model}, TAVID consists of two key pipelines: interactive video generation and conversational speech synthesis.
These two streams are synergistically interconnected through the proposed cross-modal mappers--the \textit{Motion Mapper} and the \textit{Speaker Mapper}--enabling synchronized and coherent audio-visual synthesis.
Our framework first converts a text dialogue $\mathbf{x}$ into multi-stream semantic token sequences $\mathbf{S} = [s_{\text{1}}, s_{\text{2}}]$, where each stream represents one participating speaker. 
These tokens serve as a shared representation that conditions both the video and speech generation in parallel.

\paragraph{Video Generation.}
Our video generation pipeline builds upon Hallo2~\cite{cui2024hallo2} and consists of a variational auto-encoder (VAE)~\cite{kingma2013auto}, a UNet-based visual denoiser $\epsilon_\theta$, and a parallel ReferenceNet~\cite{hu2024animate,cui2024hallo2}.
Following the latent diffusion formulation~\cite{Rombach_2022_CVPR}, the visual denoiser iteratively refines the latent variable $z_t$ at each timestep $t$, conditioned on a set of guidance signals $\mathbf{C}$ via cross-attention.
The conditioning signals $\mathbf{C}$ include reference spatial features $c_\text{ref}$, a face embedding $c_\text{face}$, and interactive motion features $c_\text{mot}$, each focusing on distinct aspects of visual reconstruction.
Spatial features from ReferenceNet enhance global coherence and the face embedding from a pre-trained face encoder~\cite{deng2019arcface} preserves portrait identity.
Notably, the interactive motion features serve as the key driver for natural dyadic interactions.
These motion features are produced by the \textit{Motion Mapper} from multi-stream semantic tokens $\mathbf{S}$.
Motivated by the strong correlations between prosody and facial dynamics~\cite{cong2023learning,kim2025faces}, we adopt prosody-aware semantic tokens\footnote{{\url{https://github.com/facebookresearch/seamless_communication/tree/main/src/seamless_communication/cli/m4t/audio_to_units}}}~\cite{kim2024paralinguistics, seamless2025joint} and validate their effectiveness in Sec.~\ref{subsec:ablation}.
The \textit{Motion Mapper} is jointly optimized with the visual denoiser $\epsilon_\theta$ using the standard diffusion objective:
\begin{equation} 
\mathcal{L}_\text{visual} = \mathbb{E}\left[ || \epsilon - \epsilon_\theta(z_t, t, \mathbf{C}) ||^2 \right], \label{eq:diffusion_loss} 
\end{equation}
where $\epsilon$ denotes Gaussian noise.

\paragraph{Speech Generation.}
The speech generation pipeline consists of text-to-semantic module, acoustic denoiser and a pre-trained vocoder~\cite{lee2022bigvgan}.
Following CoVoMix~\cite{zhang2024covomix}, we employ an encoder–decoder architecture for the text-to-semantic module and flow matching algorithm~\cite{lipman2022flow} for the acoustic denoiser. 
The text-to-semantic module takes a text dialogue, tokenizes it with a BERT tokenizer~\cite{devlin2018bert, hayashi2019pre}, and predicts the multi-stream semantic tokens $\mathbf{S}$.
This module is optimized with a cross-entropy loss as follows:
\begin{equation}
    \mathcal{L}_\text{text2semantic} = - \mathbb{E} \Big[ \sum_i \log P_\theta(s_j^i \mid s_j^{<i}, \mathbf{x}) \Big],
    \label{eq:text2semantic_loss}
\end{equation}
where $\mathbf{x}$ refers to the input text dialogue and $s_{j}^i$ denotes the $i^\text{~th}$ semantic token for participant $j\in\{\text{1}, \text{2}\}$.

The subsequent acoustic denoiser converts the multi-stream semantic tokens $\mathbf{S}$ into a mixed mel-spectrogram $\mathbf{y}$.
The token streams are first embedded via a lookup table, and speaker embeddings $e_\text{spk}$ are conditioned through DSLN module~\cite{lee2022pvae} to provide voice characteristics for each stream.
In our framework, the speaker embedding can be extracted either from audio or from the face.
During training, we use audio-driven speaker embeddings extracted from the target utterance~\cite{cam++}, while at inference, we employ face-driven embeddings predicted from a reference image via the \textit{Speaker Mapper}.
Given data samples from the prior $\mathbf{y}_0 \sim p_0(\mathbf{y})$ and target distribution $\mathbf{y}_1 \sim p_1(\mathbf{y})$, the acoustic denoiser is trained to estimate the target vector field $\mathbf{y}_1 - \mathbf{y}_0$ under the flow matching objective:
\begin{equation}
    \mathcal{L}_\text{acoustic} =
    \mathbb{E}
    \Big[
    \big\|
    (\mathbf{y}_1 - \mathbf{y}_0)
    - v_t(\mathbf{y}_t \mid \mathbf{S}, e_\text{spk}^\text{audio}; \theta)
    \big\|^2
    \Big],
    \label{eq:acoustic_loss}
\end{equation}
where $\mathbf{y}_t = (1 - t)\mathbf{y}_0 + t\mathbf{y}_1$ denotes a noisy latent at timestep $t \in [0, 1]$, and $v_t(\mathbf{y}_t \mid \mathbf{S}, e_\text{spk}^\text{audio}; \theta)$ represents the estimated vector field conditioned on the semantic tokens $\mathbf{S}$ and the audio-driven speaker embedding $e_\text{spk}^\text{audio}$.

\paragraph{Cross-modal Mappers.}
While conditioning both pipelines on shared semantic tokens $\mathbf{S}$ provides a basic means of synchronization, it is insufficient to capture the rich audio-visual correlations present in real human communication.
To address this, we propose cross-modal mappers, which establish a bi-directional information pathway between video and speech generation, going beyond simple parallel generation.
To be specific, the \textit{Motion Mapper} translates multi-stream semantic tokens $\mathbf{S}$ into interactive motion features $c_\text{mot}$, including precise lip movements and responsive listening behaviors.
Conversely, the \textit{Speaker Mapper} leverages complementary visual cues from the reference image to generate vocal characteristics that closely align with visual identity.
This exchange of information enables TAVID to produce tightly coupled audio-visual outputs, constructing a synergistic loop between the audio and visual pipelines.

\subsection{Motion Mapper}
\label{sec:motionmap}
The \textit{Motion Mapper} aims to generate interactive motions from multi-stream semantic tokens $\mathbf{S}$.
This requires not only capturing the specific information of each stream but also modeling the interdependencies between the two streams.
To identify the most effective architecture for the \textit{Motion Mapper}, we explore four different strategies.
The first strategy simply adds the two streams, $s_{\text{1}}$ and $s_{\text{2}}$, before passing them to the projection layer. 
However, in preliminary experiments, this method failed to differentiate between $s_{\text{1}}$ and $s_{\text{2}}$, preventing the model from learning proper turn-taking behavior.
The second approach concatenates the two streams and feds them into the projection layer, as shown in \cref{fig:motionmapper:concat}.
The third scheme computes motion attentions for each stream separately over video hidden states, and then adds the two attention results (\cref{fig:motionmapper:dual})\footnote{We refer to this strategy as ``Dual Attention".}.
Both methods successfully differentiate between the streams and capture appropriate transitions.
However, they are limited in capturing the inter-stream correlation, which hinders interactive video generation.

To this end, we introduce a joint self-attention mechanism, inspired by the MMDiT block~\cite{esser2024scaling}, which effectively models both modality-specific behaviors and cross-modal correlations between text and image.
As shown in \cref{fig:model}, each semantic stream is first modulated through its respective LayerNorm and Linear Layer.
The joint attention module then captures the mutual dependencies between the two stream features while preserving their distinct semantics.
The resulting representations are further refined through an additional modulation block, after which they are concatenated and processed by three Linear layers to produce the interactive motions.
As verified in Sec.~\ref{subsec:ablation}, this design effectively captures both stream-specific information and cross-stream correlations, resulting in optimal performance among the four strategies.

\subsection{Speaker Mapper}
To achieve tight alignment between acoustic and visual identity, we introduce the \textit{Speaker Mapper} which estimates vocal characteristics from reference images.
Specifically, the mapper utilizes two complementary visual features.
First, the face embedding $c_\text{face}$ extracted from the face encoder provides personalized visual characteristics, serving as the primary visual cue. 
In addition, inspired by recent findings that highlight diffusion models as powerful representation learners~\cite{dhariwal2021diffusion, clark2023text, mukhopadhyay2024text}, we incorporate spatial features $c_\text{ref}$ obtained from ReferenceNet.
ReferenceNet is trained to explicitly preserve visual consistency over frames and its hidden features offer rich discriminative cues.

As shown in \cref{fig:model} (rightmost), the intermediate ReferenceNet features $c_\text{ref}$ are projected through convolutional layers and passed to a ResNet18~\cite{he2016deep}. 
The output features are concatenated with the face embedding $c_\text{face}$, and then fed into three Linear layers to predict the audio-driven speaker embedding $e_\text{spk}^\text{audio}$.
Here, the $e_\text{spk}^\text{audio}$ is extracted from the target utterance, and the \textit{Speaker Mapper} is trained with L2 loss which can be formulated as:
\begin{equation}
    \mathcal{L}_\text{speaker}= \mathbb{E}\left[|| e_\text{spk}^\text{audio} - f_\theta(c_\text{face}, c_\text{ref})||^2\right],
\end{equation}
where $f_\theta$ represents the \textit{Speaker Mapper}.

\section{Experimental Settings}
\subsection{Datasets}
Our dataset comprises multiple video and speech corpora.
The video datasets include single role (HDTF~\cite{zhang2021flow:HDTF} and ViCo~\cite{zhou2022responsive:VICO}) and conversational datasets (subsets of the Seamless Interaction~\cite{seamless_interaction}), where each training pair consists of \verb|<|dual-channel semantic tokens, video frames\verb|>|.
Due to single role data scarcity, we further extract single role segments from the conversational data by detecting 
active speech intervals using a pre-trained VAD~\cite{SileroVAD}.
These video datasets cover diverse individuals across a wide range of topics, spanning approximately 500 hours of video data.

For the speech corpora, we construct the dataset using single role (LibriTTS-R~\cite{koizumi2023libritts}) and conversational datasets (DailyTalk~\cite{lee2023dailytalk}, Fisher~\cite{cieri2004fisher}, and Seamless Interaction~\cite{seamless_interaction}).
Each training sample consists of \verb|<|text, dual-channel semantic tokens, mixed mel-spectrogram\verb|>|, and single role segments are additionally extracted from the conversational datasets following CoVoMix~\cite{zhang2024covomix}.
In total, the speech data comprise approximately 2k hours.

\subsection{Preprocessing}
All videos are standardized to 25 fps and resized to a resolution of 512 $\times$ 512.
For the Seamless Interaction dataset, which contains naturalistic full-body videos, face sequences are cropped using RetinaFace~\cite{deng2020retinaface}, and frames with occlusions, excessive movements, or head rotations exceeding $30^{\circ}$ are excluded for stable training.
The speech sampling rate is set to 16 kHz\footnote{For the Fisher dataset, the sampling rate is upsampled from 8 kHz to 16 kHz using a pre-trained EDNet~\cite{kwak2025ednet}.}, and acoustic noise is filtered using a pre-trained EDNet~\cite{kwak2025ednet}.
For the semantic tokens, we extract continuous features from the 35$^{\text{th}}$ layer of XLS-R~\cite{babu2022xls} and discretize them into 10k clusters using the K-means algorithm.
For single role samples, indices corresponding to silence are appended to form dual-channel semantic tokens.
80 bins mel-spectrograms are computed with a window size of 1280 and a hop size of 320, and text transcriptions are processed following CoVoMix~\cite{zhang2024covomix}.

\begin{table*}[ht]
    \caption{
    Quality comparison of \textbf{interactive head generation} on the Seamless Interaction test set. Subjective evaluation results are presented with 95\% confidence intervals. Arrows indicate the preferable direction and bold values denote the best result.
    }
    \centering
    \resizebox{\textwidth}{!}
    {
    \begin{tabular}{lcccccccccccc}
    \toprule
    \multirow{2}{*}{{Method}} & \multirow{2}{*}{Source} 
    & \multicolumn{3}{c}{Subjective Metrics} 
    & \multicolumn{8}{c}{Objective Metrics} \\
    \cmidrule(lr){3-5} \cmidrule(lr){6-13}
    & & Visual Quality$\uparrow$ & Lip Sync$\uparrow$ & Turn-taking$\uparrow$ 
    & FID$\downarrow$ & FVD$\downarrow$ 
    & LPIPS$\downarrow$ &LSE-C$\uparrow$ &LSE-D$\downarrow$
    & RPCC$\downarrow$ & {$\Delta$SID}$\downarrow$ & {$\Delta$Var}$\downarrow$ \\ 
    \midrule
    DIM & Audio 
    & 2.09$\pm$0.20 
    & 2.00$\pm$0.21 
    & 2.33$\pm$0.23  
    & 29.715 & 279.673 
    & 0.240 & 2.620 & 11.366 & 0.086 & 0.902 & 0.110 \\ 
    DIM & TTS 
    & 2.11$\pm$0.23 
    & 2.33$\pm$0.24 
    & 2.38$\pm$0.22 
    & 30.423 & 283.056 
    & 0.255 & 2.480 & 11.620 & 0.061 & 1.010 & 0.229 \\ 
    \midrule
    {\bf Ours} & Text 
    & \textbf{3.75}$\pm$\textbf{0.23} 
    & \textbf{3.80}$\pm$\textbf{0.23}
    & \textbf{3.84}$\pm$\textbf{0.24}
    & \textbf{16.625} & \textbf{179.305} 
    & \textbf{0.056} & \textbf{6.457} & \textbf{8.403}
    & \textbf{0.031} & \textbf{0.489} & \textbf{0.011} \\ 
    \bottomrule
    \end{tabular}
    }
    \label{tab:sota_interactive}
\end{table*}

\begin{figure*}[t]
    \centering
    \includegraphics[width=\textwidth]{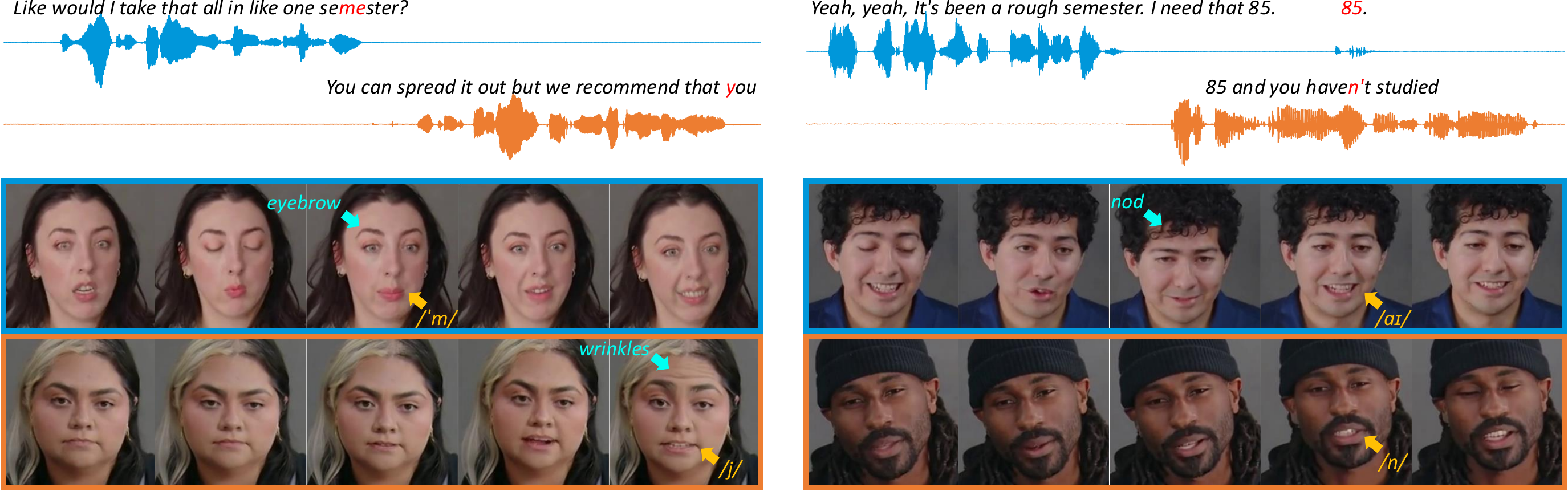}
    \caption{
    Qualitative results of TAVID on the Seamless Interaction test set. 
    Our method converts text inputs into realistic conversational speech and interactive video, capturing precise lip motions, responsive behaviors, and speech overlaps.
    }
    \label{fig:qualitative}
\end{figure*}

\subsection{Implementation Details}
In the \textit{Motion Mapper}, semantic tokens, represented as K-means cluster indices, are first transformed into continuous feature space using their corresponding original cluster centroids. 
All training is conducted on 8 NVIDIA A6000 GPUs. 
The video generation pipeline is trained for 32k steps with a batch size of 32, initializing weights from Hallo2~\cite{cui2024hallo2}, except for the \textit{Motion Mapper} and motion attention. 
At each step, 14 consecutive frames are randomly sampled, and only the weights of the \textit{Motion Mapper}, motion attention, and temporal attention are updated. 
To improve video quality, classifier-free guidance (CFG)~\cite{ho2021classifier} is applied,  dropping a reference image, semantic tokens, and motion frames with a probability of 0.05 during training.

The text-to-semantic module is trained for 250k steps with a batch size of 192, while the acoustic denoiser is trained for 320k steps with a batch size of 96. 
CFG is used to enhance speech quality. 
During training, text inputs are dropped in the text-to-semantic module with a probability of 0.1, and semantic tokens together with speaker embeddings are omitted in the acoustic denoiser with a probability of 0.3. 
In particular, we employ CFG-filter~\cite{darefsky2024parakeet} in the text-to-semantic module to mitigate the speed-drift issue observed when using vanilla CFG.
The \textit{Speaker Mapper} is trained for 1k steps with a batch size of 64.

\subsection{Evaluation Metrics}

\noindent\textbf{Video Metrics.}
We evaluate the quality of generated videos using both subjective and objective metrics. 
For subjective evaluation, we conduct 5-scale Mean Opinion Score (MOS) tests, in which 20 participants rate 20 interactive video samples based on visual quality, lip-sync accuracy, and turn-taking naturalness.
Our objective evaluation assesses overall visual fidelity and temporal realism using a suite of metrics including FID, FVD, LPIPS~\cite{zhang2018unreasonable:LPIPS}.
Audio-visual synchronization is quantified using SyncNet scores (LSE-C and LSE-D)~\cite{chung2016out,prajwal2020lip}. 
In addition, we utilize residual Pearson correlation coefficients (RPCC) to assess speaker-listener coordination, while motion diversity is evaluated using the absolute difference in pose variance ($\Delta \text{Var}$) and the SI~\cite{ng2022learning} for diversity ($\Delta \text{SID}$) with respect to the ground truth distribution~\cite{ng2023can}.

\noindent\textbf{Speech Metrics.} 
The quality of generated speech is evaluated using both subjective and objective metrics. 
For subjective evaluation, we conduct MOS tests, where 20 domain-experts evaluate naturalness and face matching score, which measures how well the generated voice matches the reference image. 
For objective metrics, we report UTMOS~\cite{saeki2022utmos} for speech quality and VoxSim~\cite{ahn2024voxsim} for speaker similarity to the target speaker. 

\section{Experimental Results}
In this section, we demonstrate the effectiveness of our approach across multiple dimensions, including interactive head generation, single role head generation, and face-stylized speech generation.
We additionally conduct comprehensive ablation studies to assess the effect of each component within TAVID.

\subsection{Interactive Head Generation}
\paragraph{Qualitative Evaluation.}
To intuitively demonstrate the generation quality of TAVID, we present two conversational video results in \cref{fig:qualitative}.
Both examples exhibit accurate lip synchronization (yellow arrows), expressive non-verbal motions (sky-blue arrows), and smooth transitions between speaking and listening states.
Particularly, the left example illustrates prosodic emphasis through raised eyebrows (top) and forehead wrinkles (bottom), highlighting dynamic facial expressions that align with prosodic variations.
In addition, the right example demonstrates that our method can generate overlaps that are both acoustically and visually natural.
Notably, despite receiving only text dialogue as input, TAVID generates both overlapped video and speech with appropriate timing, fully synchronized with each other.

\paragraph{Quantitative Evaluation.}
We perform a quantitative comparison for interactive head generation on the Seamless Interaction test set.
Since our method is the first attempt at text-driven interactive head generation, we compare our model with the recent audio-driven approach, DIM~\cite{tran2024dim}, which is the only publicly available method\footnote{\url{https://github.com/Boese0601/Dyadic-Interaction-Modeling}}.
We additionally compare a TTS-cascaded version of DIM, where the source audio is generated by our system.
To ensure a fair comparison, DIM is trained from scratch on the same training dataset as TAVID, and the results are shown in \cref{tab:sota_interactive}.

As shown, TAVID achieves high-quality video generation, significantly outperforming all the baselines across every aspect.
In particular, subjective evaluation results confirm that TAVID highly improves perceptual quality in terms of visual quality, lip-sync accuracy, and turn-taking naturalness.
Objective evaluation further supports the effectiveness of our method, demonstrating superior performance in video quality (FID, FVD, and LPIPS), lip-sync accuracy (LSE-C and LSE-D), speaker–listener coordination (RPCC), and motion diversity ($\Delta \text{Var}$ and $\Delta \text{SID}$).
It is worth noting that, due to the inherent limitations of the cascaded system, the TTS-cascaded DIM exhibits degraded performance across all objective metrics, underscoring the advantages of our text-driven end-to-end approach.

\subsection{Single Role Head Generation}

\begin{table}[t]
    \caption{
    Quality comparison of \textbf{talking head generation} on the HDTF test set. 
    The “Source” column indicates the source used for video generation.
    }
    \label{tab:thg}
    \centering
    \resizebox{\columnwidth}{!}
    {
    \begin{tabular}{lcccccccc}
    \toprule
    {Method} &{Source} &{FID}$\downarrow$ &{FVD}$\downarrow$ &{LSE-C}$\uparrow$ & 
    {LSE-D}$\downarrow$ \\ \midrule 
    SadTalker & Audio &23.398 &328.643 &7.142 
    & 7.881\\ 
    Echomimic & Audio &19.037 &276.143 &5.929 
    &9.381 \\
    Hallo2 & Audio &18.205 &258.085 &{\bf 7.816} 
    &{\bf 7.737} \\ \midrule
    SadTalker & TTS  &23.510 &328.539 &7.227 
    &7.756 \\
    Echomimic & TTS &19.316 &283.432 &5.919 
    &9.377 \\
    Hallo2 & TTS &18.742 &250.520 &7.697 
    &7.819 \\
    \midrule
    {\bf Ours}  & Text &\textbf{16.745} &\textbf{245.493} &{7.537} 
    &8.155 \\
    \bottomrule
    \end{tabular}
    }
\end{table}

\paragraph{Talking Head Generation.}

We also compare our versatile system, with a focus on talking  and listening head generation.
For talking-head generation, 20 samples from HDTF~\cite{zhang2021flow:HDTF} are used for evaluation, and we compare our text-driven method against recent audio-driven models: SadTalker~\cite{zhang2023sadtalker}, Echomimic~\cite{chen2025echomimic}, and Hallo2~\cite{cui2024hallo2}.
As shown in \cref{tab:thg}, our method achieves the best performance in video quality, outperforming all baselines in both FID and FVD, while achieving a lip-sync score comparable to the state-of-the-art method.
Moreover, similar to interactive head generation, the TTS-cascaded baselines suffer from consistent quality degradation, whereas our approach avoids this error accumulation with providing flexible control over the spoken content.

\begin{table}[t]
    \caption{
    Quality evaluation of \textbf{listening head generation} on the ViCo test set. * indicates results reported in DIM~\cite{tran2024dim}, INFP~\cite{zhu2024infp}, and ARIG~\cite{guo2025arig}.
    }
    \label{tab:lhg}
    \centering
    \resizebox{\columnwidth}{!}
    {
    \begin{tabular}{lcccccccc}
    \toprule
    \multirow{2}{*}{{Method}} &
    \multicolumn{2}{c}{{FD$\downarrow$}} & \multicolumn{2}{c}{{RPCC$\downarrow$}} &
    \multicolumn{2}{c}{{$\Delta$SID}$\downarrow$} &
    \multicolumn{2}{c}{{$\Delta$Var}$\downarrow$} \\
    \cmidrule(lr){2-3} \cmidrule(lr){4-5} \cmidrule(lr){6-7} \cmidrule(lr){8-9}
    & Exp & Pose & Exp & Pose & Exp & Pose & Exp & Pose \\
    \midrule
    L2L*  &33.93 &0.06 & 0.06 & 0.08 &2.23 &1.35 &0.47 & \textbf{0.00} \\
    RLHG* &39.02 &0.07 & 0.08 &0.02 & 1.38 &0.84 &0.22 & \textbf{0.00} \\
    DIM*  & 23.88 & 0.06 & 0.06 & 0.03 &1.29 &1.66 &0.23 & \textbf{0.00}  \\
    INFP*  & 18.63 & 0.07 & - & - & 0.22 & 0.09 &1.53 & 0.16\\
    ARIG*  & 18.39 &0.06 &0.05 &\textbf{0.01} &\textbf{0.18} & \textbf{0.07} &1.61 & 0.15\\ \midrule
    {\bf Ours} & \textbf{16.03} & \textbf{0.04} & \textbf{0.01} & {0.02} & 0.24 & 0.34 & \textbf{0.10} & \textbf{0.00}\\
    \bottomrule
    \end{tabular}
    }
\end{table}

\paragraph{Listening Head Generation.}
The quality of listening-head generation is evaluated on the ViCo test set~\cite{zhou2022responsive:VICO} against L2L~\cite{ng2022learning}, RLHG~\cite{zhou2022responsive:VICO}, DIM~\cite{tran2024dim}, INFP~\cite{zhu2024infp}, and ARIG~\cite{guo2025arig}.
In addition to RPCC, $\Delta$SID and $\Delta$Var, we compute Frechet distance (FD) to assess motion realism.

The results in \cref{tab:lhg} demonstrate that our method consistently outperforms recent approaches in listening head generation.
TAVID achieves the best results in FD and $\Delta$Var, indicating that it generates faithful non-verbal motions that closely resemble the ground truth.
Although the $\Delta$SID of TAVID shows a slight degradation compared to ARIG~\cite{guo2025arig}, our approach achieves a comparable RPCC score, highlighting its ability to generate natural responsive behaviors.

\subsection{Face-stylized Speech Generation}
The speech generation quality of TAVID is evaluated on the VoxCeleb2~\cite{chung2018voxceleb2} test set, ensuring that there is no overlap of speakers with the training dataset.
We randomly sample 30 utterances for subjective evaluation and 100 utterances for objective evaluation, and compare our method against several recent approaches.
Specifically, we include YourTTS~\cite{casanova2022yourtts} and CoVoMix\footnote{For a fair comparison, the speaker conditioning method is replaced with the use of audio-driven speaker embedding as in YourTTS~\cite{casanova2022yourtts}.}~\cite{zhang2024covomix}, which utilize audio-driven speaker embeddings extracted from the target utterance, as well as Face-TTS~\cite{lee2023imaginary} and FVTTS~\cite{lee2024fvtts}, which are based on face-driven speaker embeddings.
As shown in \cref{tab:tts}, TAVID achieves the best naturalness scores (naturalness MOS and UTMOS), indicating that it synthesizes clear and high-quality speech.
While YourTTS and CoVoMix show higher VoxSim scores than our method, this is due to the injection of vocal characteristics directly derived from the target utterance.
Among the face-stylized methods, our approach achieves the best speaker similarity.
More importantly, TAVID outperforms all baselines, including audio-stylized methods, in face matching MOS. 
This indicates that TAVID generates vocal characteristics that are perceptually consistent with the visual identity.

\begin{table}[t]
    \caption{Results of \textbf{face-stylized speech generation} on VoxCeleb2 datasets.
    All test speaker are unseen during training.}
    \centering
    \resizebox{\columnwidth}{!}
    {
    \begin{tabular}{lcccc}
    \toprule
    \multirow{2}{*}{{Method}} 
    & \multicolumn{2}{c}{Subjective Metrics} 
    & \multicolumn{2}{c}{Objective Metrics} \\
    \cmidrule(lr){2-3} \cmidrule(lr){4-5}
    &{Naturalness}$\uparrow$ &{Face Matching}$\uparrow$  &{UTMOS}$\uparrow$  &{VoxSim}$\uparrow$ \\
    \midrule
    \multicolumn{5}{l}{\textbullet~\textit{Audio-driven Voice Synthesis}} \\
    ~~YourTTS &3.20$\pm$0.20 &3.47$\pm$0.17 &3.448  & \textbf{0.662} \\
    ~~CoVoMix &3.87$\pm$0.27 &3.40$\pm$0.25 &3.146  & 0.659 \\
    \midrule
    \multicolumn{5}{l}{\textbullet~\textit{Face-driven Voice Synthesis}} \\
    ~~Face-TTS &2.25$\pm$0.21 &2.02$\pm$0.22 &1.941  & 0.278 \\ 
    ~~FVTTS &2.72$\pm$0.22 &2.83$\pm$0.23 &2.730  & 0.336\\ 
    ~~{\bf Ours}  &{\bf 4.20}$\pm${\bf 0.20} &{\bf 3.87}$\pm${\bf 0.19} &\textbf{3.530}  &\textbf{0.380}\\ 
    \bottomrule
    \end{tabular}
    }
    \label{tab:tts}
\end{table}

\subsection{Analyses}
\label{subsec:ablation}
\paragraph{Motion Mapper Architecture.}
To investigate the optimal architecture for the \textit{Motion Mapper}, we evaluate the performance of talking and listening head generation across three different strategies (concatenation, dual attention, and joint attention), as described in Sec.~\ref{sec:motionmap}.
In \cref{tab:ablation:mm_thg}, the joint attention architecture shows superior performance in the talking head generation.
It obtains the best scores in lip synchronization and image realism, indicating its capability to translate stream-specific information into precise self-driven movements.

In addition to talking head scenario, as shown in \cref{tab:ablation:mm_lhg}, the joint attention mechanism achieves robust scores in listening head generation, indicating its ability to capture cross-stream interdependencies.
While dual attention shows slightly better performance on several expression-related metrics, joint attention delivers more consistent performance across all metrics, outperforming both alternatives in pose realism and coordinated movement.
Overall, these results of talking and listening head generation collectively validate the effectiveness of the joint attention mechanism, demonstrating its ability to capture cross-stream correlations as well as stream-specific information.

\begin{figure}[t]
    \centering
    \includegraphics[width=\linewidth]{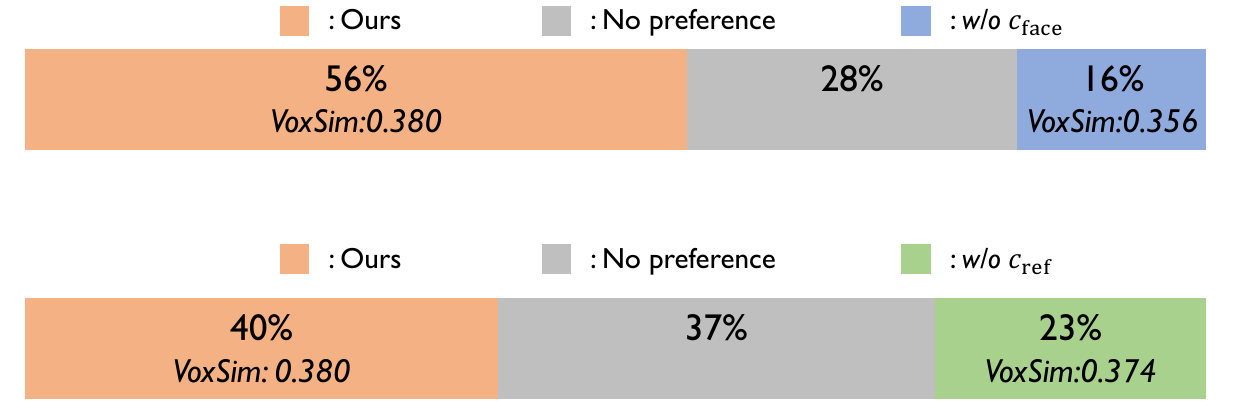}
    \caption{
    AB preference results of the \textit{Speaker Mapper} for visual and acoustic identity matching.}
    \label{fig:ablation:speakermapper}
\end{figure}

\paragraph{Prosody-aware Semantic Tokens.}

We explore the effect of prosody-aware semantic tokens (10k clusters from the hidden features of XLS-R\footnote{We denote these tokens as $\mathbf{S}_\text{prosody}$.}) by comparing our model against a variant using HuBERT~\cite{hsu2021hubert}-based semantic tokens\footnote{\url{https://github.com/vivian556123/NeurIPS2024-CoVoMix/tree/main/fairseq-hubert}}, which are known to contain rich linguistic but limited prosodic cues~\cite{polyak2021speech,chang2022distilhubert,lin2023utility}.
As shown in \cref{tab:ablation:mm_thg}, omitting prosody information degrades overall quality in talking head generation.
Specifically, lip-sync accuracy drops with a notable margin, highlighting the importance of prosodic cues for generating accurate and expressive lip motions.

The results of listening head generation consolidate the importance of prosodic cues.
In \cref{tab:ablation:mm_lhg}, the absence of the prosodic cues results in quality degradation in motion realism (FD) and natural coordination (RPCC).
These findings support our hypothesis that capturing fine-grained prosody as well as linguistics, is crucial for synthesizing realistic and interactive facial dynamics.

\paragraph{Speaker Mapper.}
To verify the effect of each visual feature used in the \textit{Speaker Mapper}, we compute VoxSim (objective) alongside AB preference tests (subjective), in which 20 subjects are asked to compare face matching scores.
The results in \cref{fig:ablation:speakermapper} validate that each feature provides a distinct contribution to enhancing face matching accuracy.
We observe that removing the face embedding $c_\text{face}$ leads to substantial degradation in both objective and subjective performance (top), confirming its central role in predicting face-matching vocal characteristics.
The absence of the ReferenceNet feature $c_\text{ref}$ results in a relatively small but notable degradation, demonstrating that it provides complementary visual cues that help align visual and acoustic identity.

\begin{table}[t]
    \caption{\textit{Motion Mapper} analysis of talking head generation.}
    \vspace{-1.5mm}
    \centering
    \resizebox{0.85\columnwidth}{!}
    {
    \begin{tabular}{lccccccc}
    \toprule
    {Method}  &{FID}$\downarrow$ &{FVD}$\downarrow$ &{LSE-C}$\uparrow$ &{LSE-D}$\downarrow$\\ \midrule
    {Concat} &17.921 &244.185 &7.296 &8.236\\
    {Dual Att.}  &17.451 &\textbf{237.997} &7.117 &8.295\\
    {Joint Att.}  &\textbf{16.745} &245.493 &\textbf{7.537} &\textbf{8.155}\\ \midrule
    ~~\textit{w/o} $\mathbf{S}_\text{prosody}$  &17.407 &266.942 &6.645 &8.924\\ 
    \bottomrule
    \end{tabular}
    }
    \label{tab:ablation:mm_thg}
\end{table}

\begin{table}[t]
    \caption{\textit{Motion Mapper} analysis of listening head generation.}
    \vspace{-1.5mm}
    \centering
    \resizebox{\columnwidth}{!}{
    \begin{tabular}{lcccccccc}
    \toprule
    \multirow{2}{*}{{Method}} &
    \multicolumn{2}{c}{{FD}$\downarrow$} &
    \multicolumn{2}{c}{{RPCC}$\downarrow$} &
    \multicolumn{2}{c}{{$\Delta$SID}$\downarrow$} &
    \multicolumn{2}{c}{{$\Delta$Var}$\downarrow$} \\
    \cmidrule(lr){2-3} \cmidrule(lr){4-5} \cmidrule(lr){6-7} \cmidrule(lr){8-9}
    & Exp & Pose & Exp & Pose & Exp & Pose & Exp & Pose \\
    \midrule
    {Concat} & 17.55 & 0.05 & 0.01 & 0.03 & 0.25 & \textbf{0.33} & 0.07 & \textbf{0.00} \\
    {Dual Att.} & \textbf{15.27} & 0.05 & \textbf{0.00} & 0.03 & 0.27 & 0.40 & 0.08 & \textbf{0.00} \\
    {Joint Att.}  & {16.03} & \textbf{0.04} & {0.01} & \textbf{0.02} & 0.24 & 0.34 & 0.10 & \textbf{0.00}\\ \midrule
    ~~\textit{w/o} $\mathbf{S}_\text{prosody}$ & 16.39 & 0.05 & 0.01 & 0.04 & \textbf{0.19} & 0.38 & \textbf{0.05} & \textbf{0.00} \\ 
    \bottomrule
    \end{tabular}
    }
    \vspace{-3mm}
    \label{tab:ablation:mm_lhg}
\end{table}

\section{Conclusion}
In this work, we presented TAVID, a unified framework for jointly generating interactive faces and conversational speech in a synchronized manner. 
To achieve precise audio-visual alignment, we proposed two cross-modal mappers which facilitate the effective cross-modal exchange of mutually complementary information and seamless integration of video and speech generation.
Extensive experiments demonstrate that our approach consistently outperforms prior methods in talking head realism, listening head responsiveness, dyadic interaction fluency, and speech quality. 
These results highlight the potential of TAVID as a versatile solution for building human-like conversational systems and underscore the importance of modeling tightly coupled audio-visual interactions in multimodal synthesis.

\clearpage
{
    \small
    \bibliographystyle{ieeenat_fullname}
    \bibliography{main}
}


\end{document}